\documentclass{article}

\usepackage{natbib}
\usepackage[preprint]{tmlr}

\usepackage[utf8]{inputenc} 
\usepackage[T1]{fontenc}    
\usepackage{hyperref}       
\usepackage{url}            
\usepackage{booktabs}       
\usepackage{amsfonts}       
\usepackage{nicefrac}       
\usepackage{microtype}      
\usepackage{xcolor}         
\usepackage{todonotes}
\usepackage{algorithm}
\usepackage{algorithmic}
\usepackage{multirow}
\usepackage{xspace}
\usepackage{comment}
\usepackage{subcaption}
\usepackage{enumitem}
\usepackage{wrapfig}

\excludecomment{nextversion}

\newcommand{\rholoss}{RHOS-Loss\xspace}

\title{The Benefits and Risks of Transductive Approaches for AI Fairness}

\author{%
  \name Muhammed T. Razzak \email muhammed.razzak@cs.ox.ac.uk \\
  \addr OATML\\
  University of Oxford
  \AND
  \name Andreas Kirsch \\
  \addr University of Oxford
  \AND
  \name Yarin Gal \\
  \addr OATML\\
  University of Oxford
}

\begin{document}

\maketitle

\begin{abstract}
Recently, transductive learning methods, which leverage holdout sets during training, have gained popularity for their potential to improve speed, accuracy, and fairness in machine learning models. Despite this, the composition of the holdout set itself, particularly the balance of sensitive sub-groups, has been largely overlooked. Our experiments on CIFAR and CelebA datasets show that compositional changes in the holdout set can substantially influence fairness metrics. Imbalanced holdout sets exacerbate existing disparities, while balanced holdouts can mitigate issues introduced by imbalanced training data. These findings underline the necessity of constructing holdout sets that are both diverse and representative.
\end{abstract}

\section{Introduction}
Algorithmic decision-making using machine learning models has seen widespread adoption in various applications that significantly impact society, including healthcare, finance, criminal justice, and employment. However, when these machine learning models are unfair or biased, their predictions can lead to severe adverse effects on individuals and communities~\citep{buolamwini18}. Such biases can perpetuate and even exacerbate existing inequalities, causing harm to marginalized groups and undermining public trust in these systems~\citep{GUEGAN18,buolamwini18}. Addressing and mitigating bias in machine learning models is therefore crucial to ensure that these technologies are deployed ethically and equitably.

Transductive learning approaches, which use a holdout set to guide training, are gaining popularity and are an area of active research.
The goals of these approaches vary: they range from improving generalization and robustness \citep{ren2018learning, saxena2019data, shu2019meta, vyas2020learning}, class imbalance \citep{lin2017focal, kumar2010self, dong2017class}, or reducing training time \citep{mindermann2021} to achieving convergence by learning a curriculum \citep{kumar2010self, saxena2019data}.
Importantly, some methods use a holdout set to guide training to encourage fairer outcomes \citep{zhao2019metric,jiang2018mentornet,shu2019,seto2022}. 

Crucially, these methods implicitly assume that the holdout set used is unbiased or fair; an assumption that may not hold in practice and which could degrade performance. 
Despite this strong assumption, the robustness of these methods to changes in the composition of the holdout set has not been examined. 
At the same time, these methods could be particularly vulnerable to inheriting and magnifying the biases present in holdout sets and induce generalization failures beyond the holdout sets.

In this work, we address this potentially critical oversight by investigating how the composition of the holdout set, in terms of balance across sensitive sub-groups, impacts model fairness.
We examine both the benefits and risks using the recently published \rholoss \citep{mindermann2021} and FairGen~\citep{choi20}, through controlled experiments on image datasets CIFAR100-20\footnote{A variant of CIFAR100 where we use the 20 defined superclasses as targets and the regular finegrained classes as protected information.}~\citep{cifar} and CelebA~\citep{celeba}.

We find evidence that even less severe compositional changes to the holdout set significantly affect fairness metrics such as equal opportunity or predictive equality.
Our results reinforce the need for carefully constructing diverse and representative holdout sets that closely align with the deployment goals when applying transductive learning.
This is of particular importance as data sub-sampling and automated filtering become more prominent, and machine learning models become larger and harder to evaluate.

Using biased holdout sets could have severe effects in sensitive domains: for example, if a company were to use transductive machine learning for screening job applicants' resumes but unknowingly uses a holdout set that under-represents certain demographic groups, the resulting model might systematically discriminate against them, reinforcing societal inequities---even if the available training data had sufficient coverage for these groups; likewise, in healthcare, if a model for disease diagnosis were trained on data transductively selected with a badly curated holdout set, it might fail to detect conditions that predominantly affect underrepresented patient populations, leading to poorer health outcomes.

\paragraph{Outline} The paper is structured as follows:
\begin{enumerate}[noitemsep]
    \item In section \ref{sec:related}, we examine related work for data balancing and fairness within generative settings;
    \item In section \ref{sec:background}, we look at two recent transductive training methods;
    \item In section \ref{sec:experiments}, we empirically study the effects of holdout set composition on performance using CIFAR and CelebA. Specifically, we demonstrate that changes to the balance of sensitive sub-groups in the holdout set can significantly impact model fairness metrics like equal opportunity and predictive equality, in both discriminative settings (section \ref{sec:disc} and generative settings (section \ref{sec:gen}), and hypothesize on the reasons (section \ref{sec:discussion});
    \item Finally, in section \ref{sec:conclusion}, we discuss the importance of our study on transductive model fairness and suggest directions for future work.
\end{enumerate}

\section{Related Work}
\label{sec:related}

Existing approaches address fairness both in the discriminative and generative setting. The most relevant for our work are fairness reweighting methods and the literature about fairness in generative models.
 
\paragraph{Fairness Reweighting Methods}
Data reweighting is a common approach to mitigate fairness issues and is effective at dealing with class imbalances. Classical approaches to data reweighting consist of resampling data \citep{kahn1953methods,romano2020achieving}, using domain-specific knowledge \citep{zadrozny2004learning}, estimating weights based on data difficulty \citep{lin2017focal, malisiewicz2011ensemble}, and using class-count information \citep{cui2019class, roh2021sample}. Weighting based on fairness metrics has also been explored \citep{jiang2018mentornet,zhao2019metric,jiang2020identifying, seto2022}.


\paragraph{Fairness in Generative Models} %
Fairness within the context of generative models (specifically generative adversarial models), has been studied and improved since they garnered success in generating somewhat realistic images. \citet{choi20} attempts to increase fairness in GANs given the sensitive attributes. \citet{otherfairgan} targets and improves the fairness of generated datasets on sensitive attributed without knowing them a-priori. Subsequent works~\citep{biasgan, fairgen, uncertain} have built upon this. In our work here, we focus primarily on studying the effect of holdout sets on fairness in transductive models rather than improving.

\section{Background}
\label{sec:background}

We will examine two different transductive training methods in our empirical study, one each for the discriminative and generative setting. In this section, we discuss both and the fairness metrics we will use for evaluation next.

\subsection{Transductive Training Methods}

Transductive learning methods leverage unlabeled holdout or validation data during training to improve generalization performance on similar data. 

In transductive learning, the goal often is not to learn a model that generalizes to arbitrary new test data but to specifically improve performance on the given holdout set at hand, where performance is loosely defined and can refer to model training time, accuracy and fairness in the past, among others, depending on the work.

Some recent transductive learning methods include the reducible holdout loss (RHO-Loss)~\citep{mindermann2021}, FORML~\citep{seto2022}, FairGen~\citep{fairgen}, and the diversify and disambiguate~\citep{lee2023diversify}. These methods leverage additional unlabeled data from the holdout distribution during training. We examine two methods: RHO-Loss and FairGen. RHO-Loss trains a model to prioritize training on points that most reduce the loss on a provided holdout set, while FairGen attempts to train a more fair generative model by matching a fair holdout/reference distribution. Both methods use transductive learning, but for different objectives. 
\begin{wrapfigure}{r}{8cm}
    \centering
    \includegraphics[width=\linewidth]{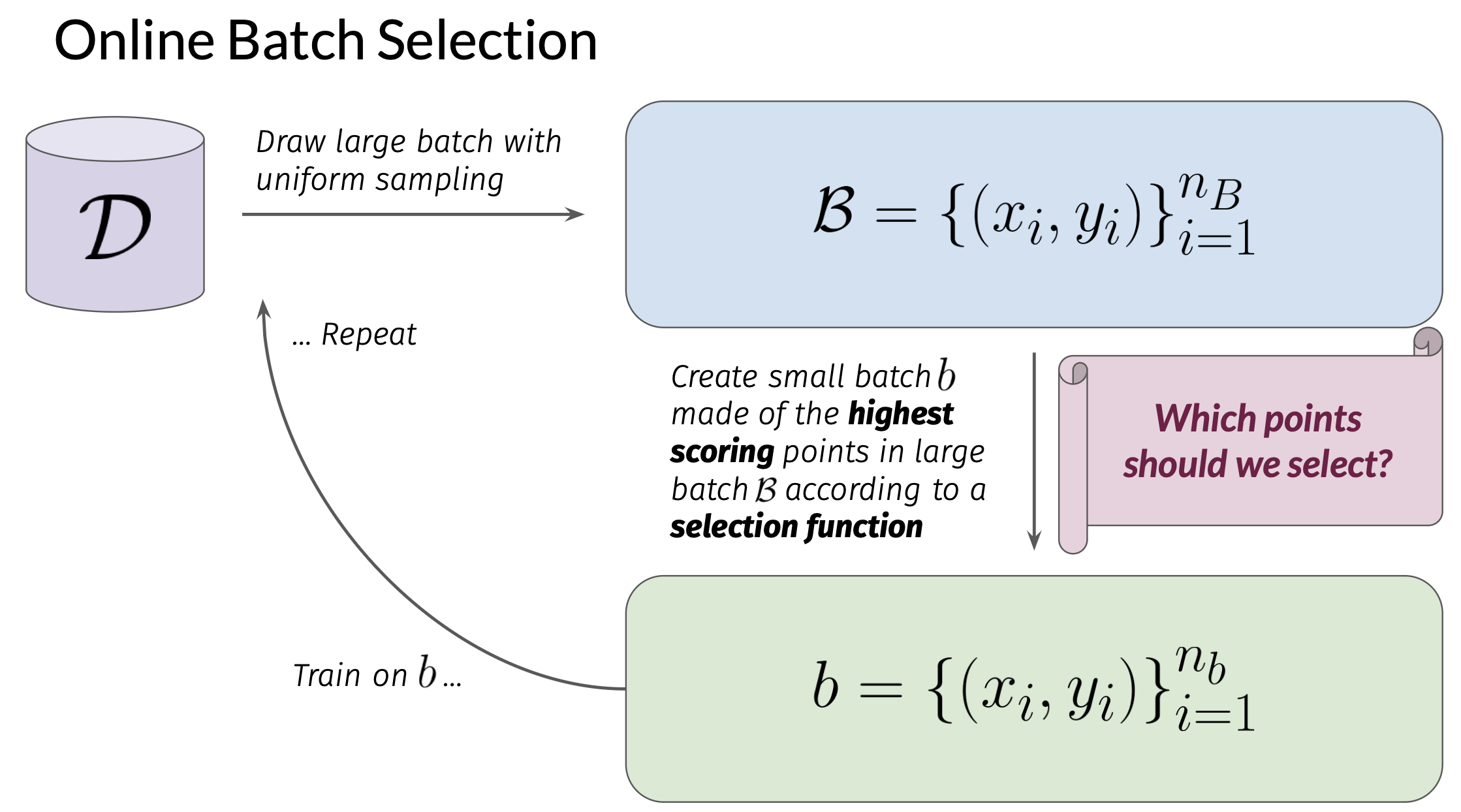}
    \caption{Illustration of the online batch setting set-up as used in \citet{mindermann2021}.}
    \label{fig:obs}
\end{wrapfigure}

\paragraph{Reducible Holdout Loss (RHO-Loss)}
The reducible holdout loss (RHO-Loss) method~\citep{mindermann2021} provides an implementation of transductive learning in an online batch selection setting (figure \ref{fig:obs}). They use transductive learning to improve the speed of training large models. In online batch selection, one draws a large batch and uses a selection function to select points for a smaller batch on which to actually train a model. RHO-Loss computes two losses for each candidate training point $(x,y)$ in the large batch $B$: 1) the training loss on the model being trained, and 2) the "irreducible loss" on a model trained only on the holdout set $D_{ho}$. It prioritizes points that maximize the reducible loss, which is the difference between these two losses. This focuses training on non-redundant points worth learning for $D_{ho}$. \citet{mindermann2021} uniformly sample from the training dataset to obtain the holdout set, and they did not study the effect of the holdout set on fairness metrics as the primary objective is to reduce model training time.

\paragraph{FairGen}
One example of an approach that takes into account fairness in the generative setting is FairGen \citep{choi20}.  FairGen trains a generator $G$ on a biased dataset $D_{bias}$, but reweights the losses to attempt to match a reference distribution $p_{ref}(x)$. 

To correct for biases present in the larger $D_{bias}$ with respect some reference dataset $D_{ref}$, FairGen trains a classifier $c(y|x)$ to estimate density ratios $w(x) = \frac{p_{ref}(x)}{p_{bias}(x)}$. The generator $G$ is then trained with importance weighting on $D_{bias}$:

$$
L(G) = \mathbb{E}_{x \sim p_{bias}} [w(x) \ell(G(z), x)]
$$

where $w(x)$ upweights underrepresented points and downweights overrepresented points from $D_{bias}$.

This allows FairGen to leverage a reference dataset to train $G$ to match $p_{ref}(x)$. The key advantage is that while collecting a large representative reference dataset is difficult, a reasonably fair generator can be trained with a modest amount of reference data.

However, what happens if a fairly represented reference dataset is not obtained? Or what if a dataset is fairly represented with respect to one set of attributes, but not another. We believe this a common pathology in these methods, and if applied in the real-world can have underappreciated consequences.

\subsection{Fairness Metrics}

Depending on the setting, we use different metrics to measure \textit{equal opportunity} and \textit{predictive equality}.

\paragraph{Discriminative Setting}
We will evaluate models on their \textit{accuracy} and quality of generated images, and to assess their fairness we evaluate using the notions of \textit{equal opportunity} and \textit{predictive equality}.

\textit{Equal opportunity} requires equal true positive rate across groups:
  \begin{equation}
    P(\hat Y = 1 \mid Y=1, A=a) = P(\hat Y = 1 \mid Y=1, A=b), \forall a, b \in \mathcal A.
  \end{equation}

We measure the True Positive Rate Disparity (TPRD) (i.e., the disparity between the maximum TPR and minimum TPR) to evaluate equal opportunity.

\textit{Predictive equality} requires equal false negative rate across groups:
  \begin{equation}
    P(\hat Y = 1 \mid Y=0, A=a) = P(\hat Y = 0 \mid Y=1, A=b), \forall a, b \in \mathcal A.
  \end{equation}
    We measure the maximum false negative rate (maxFNR) evaluate predictive equality.

\paragraph{Generative Setting}
The \textit{Fréchet Inception Distance (FID)} is used, in the generative model setting, to evaluate the quality of the images generated as in standard in the field. FID leverages the Inception-v3 neural network, pretrained on large-scale image datasets such as ImageNet, to extract feature representations from both real and generated images. FID calculates the Fréchet distance, which is a measure of similarity between multivariate Gaussian distributions, based on these feature representations. FID provides an evaluation of the fidelity and diversity of generated images. To assess the fairness, we utilize a notion of \textit{predictive equality}, looking at the discrepancy between FID score of the worst performing class and the best performing class (abbreviated to \textit{FID D}). 

\section{Empirical Study}
\label{sec:experiments}
In this section, we study how the composition of the holdout sets affect the overall accuracy, generative quality and/or fairness of these models.

\paragraph{Datasets}  
We utilize CIFAR100-20 \citep{cifar} and Celeb-A \citep{celeba}. For CIFAR100-20, there is no pre-defined sensitive attribute; therefore, we use the class label as the sensitive attribute. For Celeb-A, we define the sensitive attribute to be the gender, and the objective to predict whether the individual is smiling. We use the code provided by \citet{choi20} to bias the data based on the sensitive attributes (e.g. gender and hair). CIFAR100-20 is the coarse-grain version of CIFAR100, which contains 20 coarse classes that each have 5 finegrained classes. This allows us to treat the finegrained classes as the protected attributes to evaluate fairness on. We construct the holdout set as a subset from half the training set, with the remaining half being used to train the primary model. We highlight this as the accuracy and generative quality scores will be lower than what is typically seen in papers, due to less data being used to train the model. Finally, metrics are reported over the test set, which is balanced, unless stated otherwise. 

\paragraph{Methods} 
We conduct our study using \rholoss~\citep{mindermann2021} and FairGen~\citep{choi20}. We utilize \rholoss selection \citep{mindermann2021} to study the effect of varying holdout sets on discriminative modeling, while using FairGen to study the effect of varying holdout sets on generative modeling.

\paragraph{Model Architectures} We train a ResNet-18 architecture \citep{he2016deep}, adapted for CIFAR, as the model architecture with the AdamW optimizer. We use the default PyTorch hyperparameters. For the generative modelling scenario, we used a Diffusion Model~\citep{diffusion} with a U-Net~\citep{unet} backbone. The detailed hyperparameters are listed in appendix \ref{app:model}.

\subsection{Discriminative Modeling}
\label{sec:disc}
We train two classifiers for \rholoss one on the holdout set, and we use that one to assist in training the second on the training set.

The composition of holdout set, in these transductive methods, has an exaggerated effect on the training of models. Here, we study how changes in the holdout set affect the accuracy of the models. We find that when we have a balanced holdout set compared, where the training set is imbalanced, we oversample minority groups during training, resulting in fairer model. On the other hand, an imbalanced holdout set can have a devastating impact on a model trained with a balanced training set, by undersampling minority classes. 

We ablate the effect of having a balanced, imbalanced and highly imbalanced holdout set and its effect on model training on a balanced, imbalanced and highly imbalanced in the CIFAR100-20.

\begin{table}[t]
    \caption{Fairness metrics: True Positive Rate Disparity (TPRD), Maximum False Negative Rate (maxFNR), and test accuracy for ResNet-18 on CIFAR100-20. The means and standard errors are from 5 runs. Over coarse labels.}
    \label{table:imbalance}
  \centering
  \resizebox{0.8\columnwidth}{!}{
  \begin{tabular}{lccc}
    \toprule
    Training / Holdout Set        & TPRD (\%)  $\downarrow$           & maxFNR (\%) $\downarrow$ & Accuracy (\%) $\uparrow$    \\
    \midrule
    Balanced / Highly Imbalanced    & $36.6$      & $45.8$      & $71.02$ \\
    Balanced / Imbalanced                      & $27.0$  & $41.2$ & $73.85$ \\
    Balanced / Balanced                      & $21.2$  & $37.8 $ & $80.84$ \\
    \bottomrule
  \end{tabular}
  }
\end{table}

\begin{figure}
  \centering
    \begin{subfigure}{0.34\textwidth}
    \includegraphics[width=\textwidth]{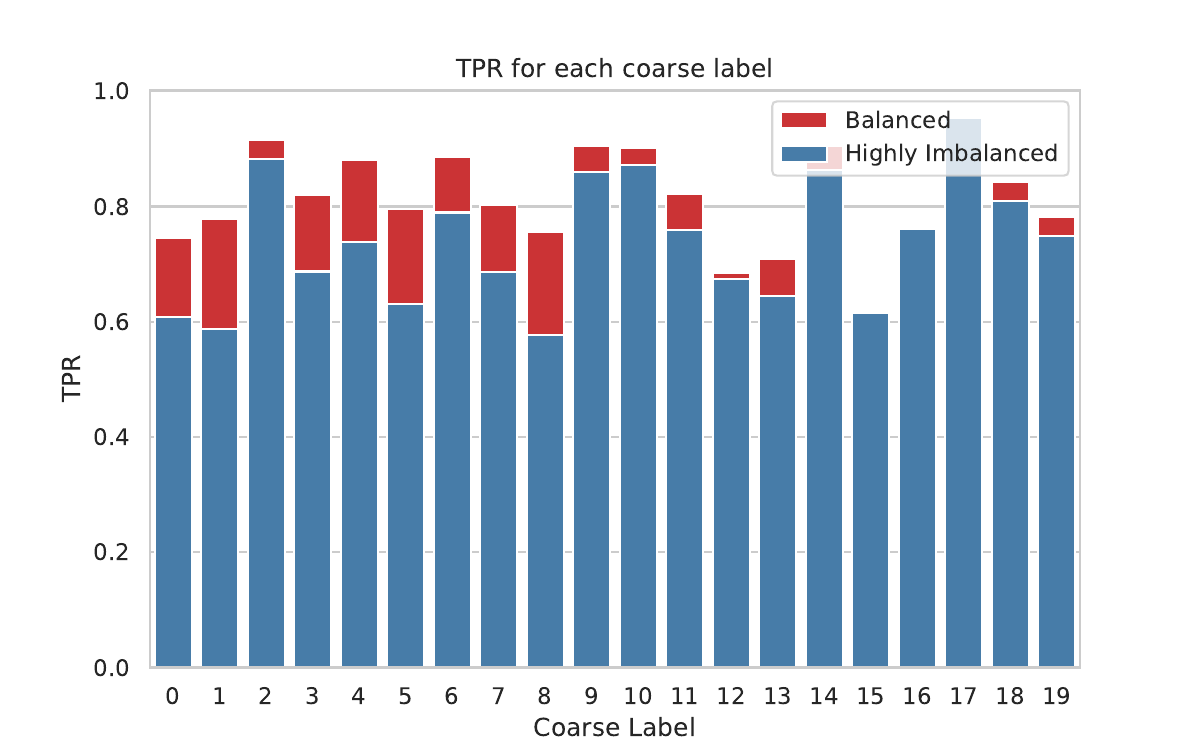}
    \caption{TPR over the coarse labels.}
    \label{fig:cifar_subsets}
    \end{subfigure}
    \hfill 
    \begin{subfigure}{0.64\textwidth}
    \includegraphics[width=\textwidth]{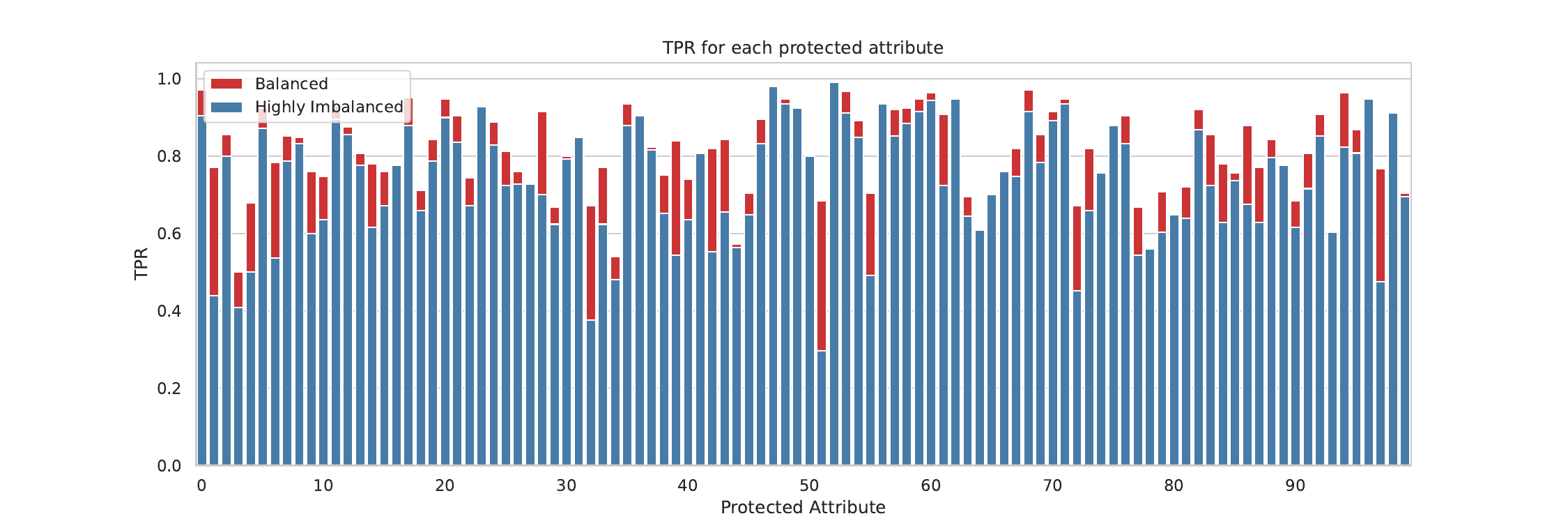}
    \caption{TPR over the protected attributes.}
    \label{fig:cifar_importance_weights}
    \end{subfigure}
  \caption{True Positive Rate of different classes and protected attributes on CIFAR20 on the balanced and highly imbalanced holdout sets.}
  \label{fig:tpr}
\end{figure}

Treating the coarse labels as the protected attribute, when the highly imbalanced holdout set is used instead of the balanced one, we see a big increase in the TPRD (table \ref{table:imbalance}). We also see big increases in overall accuracy, which could hide the deterioration of the fairness metrics.

Treating the finegrained classes within each coarse class as protected attributes, we observe in the flowers coarse class, containing orchids, poppies, roses, sunflowers, and tulips, the balanced holdout set improves true positive rates for underrepresented orchids and tulips compared to a highly imbalanced holdout. However, for overrepresented classes like roses, true positive rates are slightly reduced. In addition, while somewhat obvious, it is important to mention that the representativeness of finegrained (i.e. the sensitive attributes) assists in the overall accuracy of a coarse-grained class. This is visualised in figure \ref{fig:cifar_importance_weights}.

\subsection{Generative Modelling}
\label{sec:gen}
In this setting we use FairGen \citep{choi20} to assist in training the diffusion model with the U-Net architecture. We alter only the holdout set while keeping the training set fixed. 
\begin{table}[t]
\caption{Results of the diffusion model trained on Celeb-A and CIFAR100-20 datasets with varying degrees of bias. FID measures the similarity between generated and real images. FID D (FID Discrepancy) is the difference in FID scores between the best and worst attribute-wise FID scores.}
\centering
\label{tab:results}
\begin{tabular}{@{}llcc@{}}
\toprule
Dataset & Scenario & FID D ($\downarrow$) & FID ($\downarrow$) \\
\midrule
\multirow{3}{*}{Celeb-A} & Fair Weighting & 0.35 & 5.73 \\
 & Biased & 0.51 & 6.58 \\
 & Highly Biased & 1.08 & 7.11 \\
\midrule
\multirow{3}{*}{CIFAR100-20} & Fair Weighting & 0.47 & 7.15 \\
 & Biased & 0.88 & 8.33 \\
 & Highly Biased & 1.32 & 9.02 \\
\bottomrule
\end{tabular}
\end{table}

\begin{figure}
  \centering
    \begin{subfigure}{0.49\textwidth}
    \includegraphics[width=\textwidth]{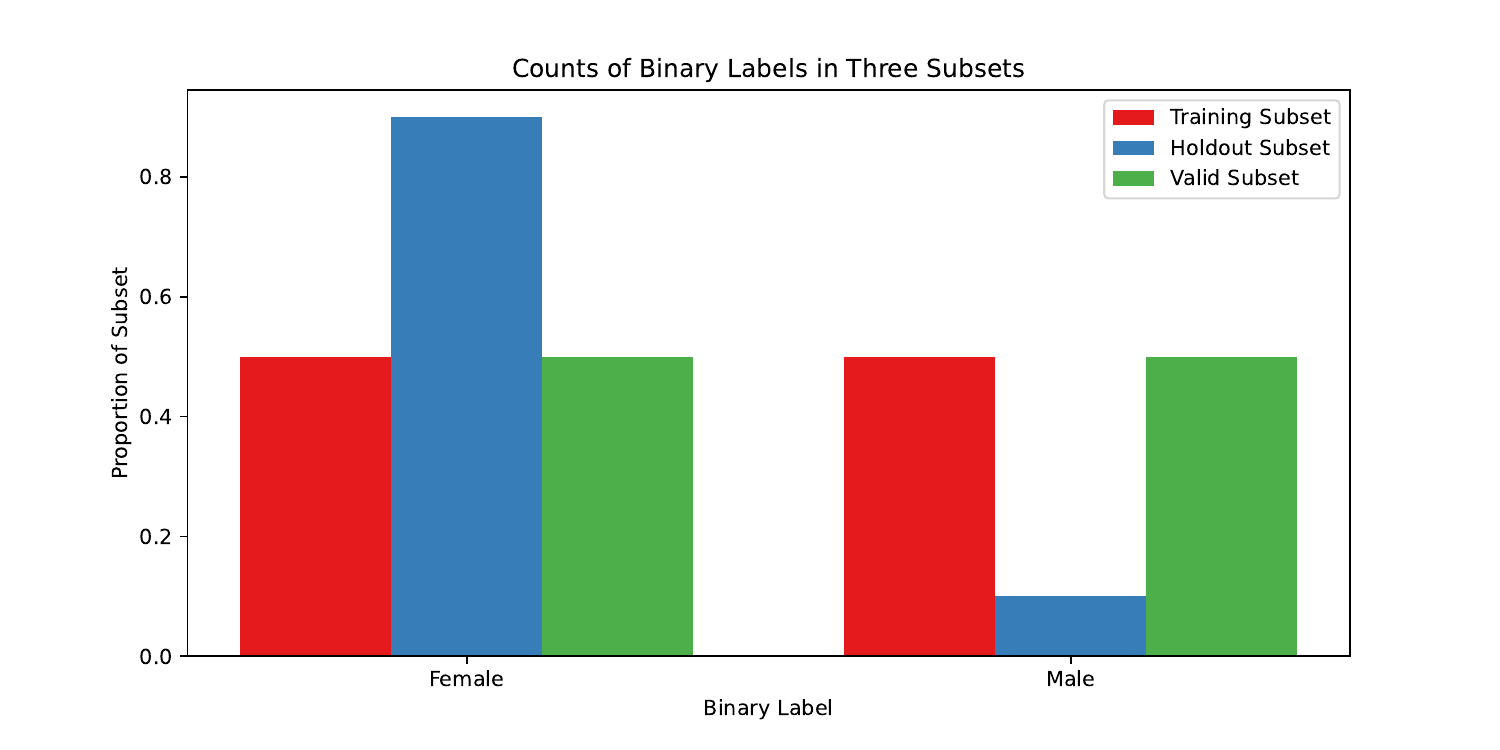}
    \caption{Dataset Subset Proportions of CelebA}
    \label{fig:celeba_subsets}
    \end{subfigure}
    \hfill 
    \begin{subfigure}{0.49\textwidth}
    \includegraphics[width=\textwidth]{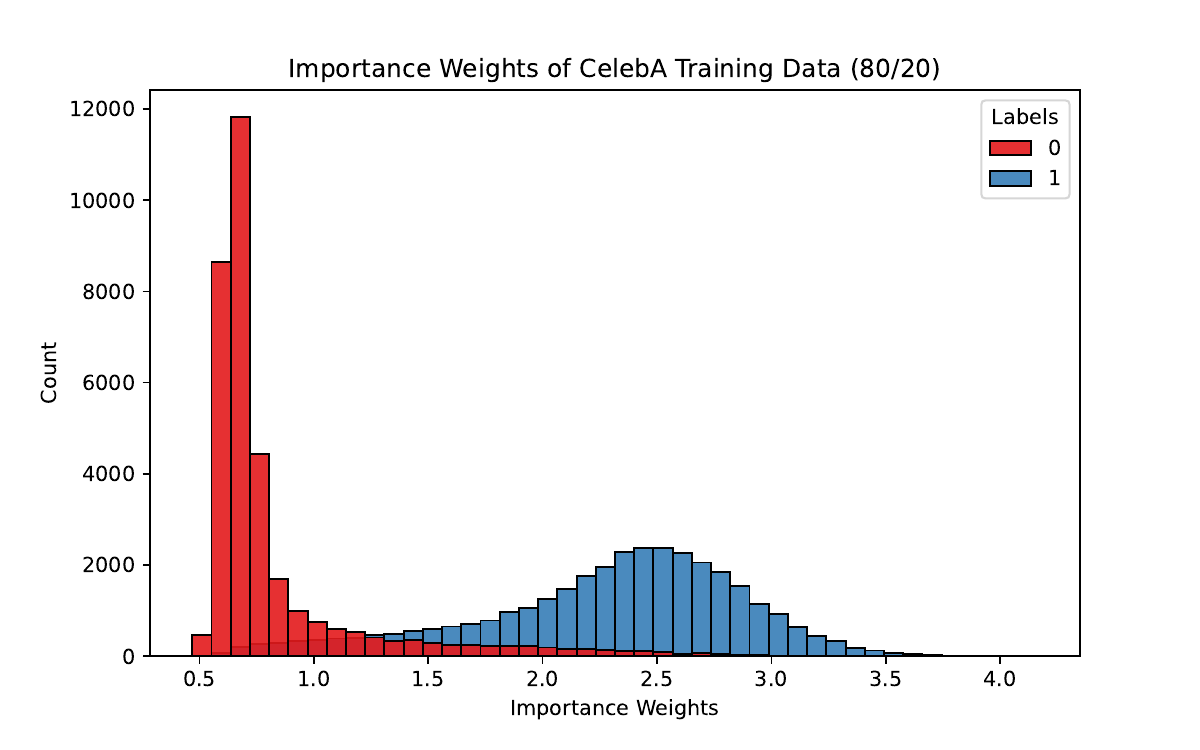}
    \caption{Importance Weights for Samples. Label 1 is male, and 0 is female.}
    \label{fig:celeba_importance_weights}
    \end{subfigure}
  \caption{The dataset proportions and importance weights for the CelebA dataset used in the generative modeling setting. Minority class (males) are downweighted and Majority class upweighted as a result of the imbalance in the reference dataset.}
  \label{fig:overall}
\end{figure}

We analyze the effect of holdout set balance on generative modeling performance and fairness on the CelebA dataset and CIFAR100-20 datasets. 
In CelebA, we test three weightings in holdout set: (1) 90\% female, 10\% male, (2)  65\% female, 35\% male (3) 50\% female, 50\% male. The training set is balanced, and we've visualized the make-up of these subsets in figure ~\ref{fig:celeba_subsets}. The resulting trained density classifier provides importance weights that naturally downweight male examples, while upweighting female examples (figure \ref{fig:celeba_importance_weights}).

Equalizing gender representation in the holdout set significantly improves FID for generated male faces from 7.2 with an imbalanced holdout to 6.17 with a balanced holdout, nearing the FID for generated females. This demonstrates that a skewed holdout hampers the model's ability to reconstruct minority features. 

On CIFAR100-20, we generate samples for each coarse class and evaluate FID for the finegrained protected attribute subgroups within each class. For the flowers class, a balanced holdout set markedly improves FID for the underrepresented orchids (from 8.9 to 8.1 from Biased to Fair) while maintaining fidelity for overrepresented roses (7.6).

The consistent FID improvements on CelebA and CIFAR100-20 protected subgroups from a balanced holdout highlight its importance for generative modeling. Imbalanced holdouts lead to lower sample quality for minority groups.

\subsection{Discussion}
\label{sec:discussion}

We hypothesize that the high sensitivity of transductive methods to holdout set imbalances can be attributed to the fact that these approaches explicitly aim to optimize performance on the holdout set: during training, the model iteratively selects training samples to minimize the loss on the holdout examples. 

If certain groups are underrepresented in the holdout set, the model will prioritize fitting well to the over-represented groups at the expense of the underrepresented ones. 
Over many training iterations, these initially small biases can compound and can substantially and adversely effects the fairness of the final model. 
Theoretically, this could be understood as a form of overfitting to the biased holdout distribution, leading to poor generalization to the true data distribution.

\section{Conclusion}
\label{sec:conclusion}
In this work, we have empirically examined how the composition of the holdout set, particularly the balance between sensitive subgroups, effects model accuracy, quality, and fairness metrics when using transductive learning methods.
Empirical evidence through controlled experiments on the CIFAR100-20 and CelebA datasets suggests that imbalances in the holdout set can lead to significant disparities in fairness metrics such as equal opportunity and predictive parity. 

Importantly, the holdout set composition can override the composition present in the training data and determine the final model's fairness as measured by different fairness metrics. 
An imbalanced holdout set can undermine fairness for models trained on balanced data, while, 
conversely, balanced holdout sets can ameliorate gaps caused by imbalanced training data, when the hold-out sets are constructed with care. This demonstrates the outsized influence of the holdout set as a representative sample guiding transductive training and how critical carefully constructing holdout sets to be representative of the target population and to include a diverse range of sensitive attributes is.

Our results reinforce the need for great care when constructing holdout sets to ensure diversity and mitigate biases that could be inherited by models otherwise. 
This highlights important open questions around developing techniques to actively construct or update holdout sets to better align with target test distributions and otherwise underpresented groups, underscoring the heightened sensitivity of transductive methods to holdout set distribution mismatch. 
Improving the robustness and fairness of transductive learning will be key to enabling the safe deployment of transductive learning approaches.

\section*{Acknowledgments}
We acknowledge the helpful feedback by Luca Zapella.

\newpage
\bibliography{fairness.bib}

\newpage
\appendix


\section{Model Hyperparameters}
\label{app:model}
\begin{table}[h]
    \centering
    \caption{Default Hyperparameters for ResNet-18}
    \label{tab:hyperparameters}
    \begin{tabular}{l l}
        \toprule
        \textbf{Hyperparameter} & \textbf{Default Value} \\
        \midrule
        Model Architecture & ResNet-18 \\
        Optimizer & AdamW \\
        Learning Rate & 0.001 \\
        Batch Size & 32 \\
        Epochs & 50 \\
        Weight Decay & 0.01 \\
        $\beta_1$ (AdamW) & 0.9 \\
        $\beta_2$ (AdamW) & 0.999 \\
        $\epsilon$ (AdamW) & $1\times10^{-8}$ \\
        \bottomrule
    \end{tabular}
\end{table}

\begin{table}[h]
    \centering
    \caption{Default Hyperparameters for a U-Net Diffusion Model}
    \label{tab:hyperparameters}
    \begin{tabular}{l l}
        \toprule
        \textbf{Hyperparameter} & \textbf{Default Value} \\
        \midrule
        Model Architecture & U-Net \\
        Optimizer & Adam \\
        Learning Rate & 0.001 \\
        Batch Size & 64 \\
        Epochs & 100 \\
        Weight Decay & 0.01 \\
        Learning Rate Schedule & Cosine Annealing \\
        Temperature & 1.0 \\
        Number of Diffusion Steps & 1000 \\
        Noise Levels & Geometric Steps \\
        \bottomrule
    \end{tabular}
\end{table}
\end{document}